\documentclass[journal]{IEEEtran}

\usepackage{times}
\usepackage{xcolor}
\usepackage{soul}
\usepackage[utf8]{inputenc}
\usepackage[small]{caption}
\usepackage{amssymb}
\usepackage{multirow}
\usepackage{epsfig}
\usepackage{graphicx}
\usepackage{amsmath}
\usepackage{amsthm}
\usepackage{color}
\usepackage{caption}
\usepackage{algorithm}
\usepackage{algorithmic}
\newcommand{\tabincell}[2]{\begin{tabular}{@{}#1@{}}#2\end{tabular}}

\ifCLASSINFOpdf
\else
\fi

\begin{document}
%
\title{Tensor graph convolutional neural network}
%
%
%

\author{Tong~Zhang,
        Wenming~Zheng,~\IEEEmembership{Member,~IEEE,}
        Zhen~Cui
        and Yang Li
\thanks{Tong Zhang and Yang Li are with the Key Laboratory of Child Development and Learning Science of Ministry of Education, and the Department
of Information Science and  Engineering, Southeast University, China. (e-mail: tongzhang@seu.edu.cn;yang\_li@seu.edu.cn).}
\thanks{Wenming Zheng is with the Key Laboratory of Child Development and Learning Science of Ministry of Education, Research Center for Learning Science, Southeast University, Nanjing, Jiangsu 210096, China (e-mail: wenming\_zheng@seu.edu.cn; xhzongyuan@seu.edu.cn). }
\thanks{Zhen Cui is with the School of Computer Science and Engineering, Nanjing University of Science and Technology, Nanjing, China. (e-mail: zhen.cui@njust.edu.cn).\protect\\
}
}

\maketitle

\begin{abstract}
In this paper, we propose a novel tensor graph convolutional neural network (TGCNN) to conduct convolution on factorizable graphs, for which here two types of problems are focused, one is sequential dynamic graphs and the other is cross-attribute graphs. Especially, we propose a graph preserving layer to memorize salient nodes of those factorized subgraphs, i.e. cross graph convolution and graph pooling. For cross graph convolution, a parameterized Kronecker sum operation is proposed to generate a conjunctive adjacency matrix characterizing the relationship between every pair of nodes across two subgraphs. Taking this operation, then general graph convolution may be efficiently performed followed by the composition of small matrices, which thus reduces high memory and computational burden. Encapsuling sequence graphs into a recursive learning, the dynamics of graphs can be efficiently encoded as well as the spatial layout of graphs. To validate the proposed TGCNN, experiments are conducted on skeleton action datasets as well as matrix completion dataset. The experiment results demonstrate that our method can achieve more competitive performance with the state-of-the-art methods.
\end{abstract}


%
\IEEEpeerreviewmaketitle

\section{Introduction}
%
%
%
%
\IEEEPARstart{G}{raph} aphs such as recommending system (the connection of users and goods). In fact, all these graphs can be easily represented as a composition of several subgraphs. In this paper, we will focus on how to perform more efficient graph convolution on this type of problem. Especially, here we refer to two classic tasks: skeleton-based action recognition and recommending system.

As a representative of dynamic sequence graphs, skeleton-based action recognition has become a hot topic of computer vision, and it draws wide attention in recent years due to its wide applications, e.g. video surveillance, games
console and robot vision. In previous literatures, various algorithms have been proposed~\cite{xia2012view,chaudhry2013bio,ofli2014sequence,vemulapalli2014human,amor2016action,huang2017riemannian,zhang2017geometric,du2015hierarchical,liu2016spatio,li2017action} to deal with skeleton data based action recognition. Some of them just focus on modeling temporal evolution while  fail to well characterize the spatial dependencies among joints. Different from these algorithms, some other literatures attempt to model spatial structure by employing structure learning algorithms such as Riemannian network~\cite{huang2017riemannian}, Lie-Net~\cite{huang2017deep}, various types of recurrent neural network (RNN)~\cite{du2015hierarchical,shahroudy2016ntu,liu2016spatio,zhang2017geometric} and graph based representation~\cite{wang2016graph}. Among them,  graph representation proposed in~\cite{wang2016graph} provides an efficient way for describing the irregular graph-structured skeleton data by modeling joints of skeleton as nodes of a graph. Further more, recent year, based on spectral graph theory, graph convolution neural network (GCNN)  is proposed in~\cite{defferrard2016convolutional,kipf2016semi,monti2017geometric} for irregular data as an alternative algorithm to CNN and shows promising performance. Although graph convolution has been used for sequence data, it more constructs a graph for each frame, whilst ignore temporal correlation. In contrast, we conduct a large spatio-temporal graph convolution by simultaneously model spatial and temporal relationship of different nodes.

Another representative of artificial intelligence application based on cross attribute factorized graphs is recommending system,  which can be formed as matrix completion problem and has been investigated in previous literatures ~\cite{kalofolias2014matrix,monti2017geometric,rao2015collaborative}. In this task, two separable graphs of different attributes, i.e. column graphs and row graphs,  are provided representing
similarity of users and items and  were shown beneficial for the performance of recommender systems. ~\cite{monti2017geometric} applies graph convolution in this task considering both users and item graphs from the Fourier transform respective and achieves the state-of-the-art performance. However, quite different from~\cite{monti2017geometric}, we propose to conduct cross graph convolution for this task aiming to optimally model the relationship between every pair of nodes from two graphs.

In this paper, taking the two representative tasks, we explore convolution on factorizable graphs. More generally, we proposed a tensor graph convolutional neural network (TGCNN) to deal with the problem. For sequential dynamic graphs, especially, we leverage recursive mechanism on consecutive tensor convolution to overcome the problem of computational explosion. Therein, a graph preserving layer is proposed to recursively optimize both previous encoded spatio-temporal graph and the successively input subgraph. In each recursive step, the salient nodes in previous subgraphs are preserved and further connected to those in current input graph, so that the graph preserving layer is able to globally model all those nodes in factorized graphs. To this end, two operations are employed in the graph preserving layer, cross graph convolution (for building cross graph relationship) and graph pooling (for nodes selecting). Specifically, for cross graph convolution, we design a novel parameterized Kronecker sum operation to learn an optimal conjunctive graph. During derivation, the property of Kronecker product is utilized so that  graph filtering can be conduced on matrices in smaller size, which thus may avoid high memory and computational costs. Following the conjunctive cross graph relationship, graph pooling is used to choose an optimal subset of salient nodes for next recursive process.
To evaluate the proposed TGCNN, we conduct experiments on two large scale action datasets named NTU RGB+D (NTU) dataset~\cite{shahroudy2016ntu} and the Large Scale Combined (LSC) dataset~\cite{zhanglarge}. Moreover, to test the generalization ability of the proposed cross graph convolution on graph structural data, we also conduct an extensive experiment on a matrix completion dataset~\cite{kalofolias2014matrix,monti2017geometric} named Synthetic ‘Netflix’ dataset. The experimental results show that our method outperforms those state-of-the-arts.

In summary, our main contributions are three folds:
\begin{enumerate}
\item we propose a novel tensor graph convolutional neural network which is able to globally learn an optimal graph from multiple factorized subgraphs through a recursive learning process.
\item we design cross graph convolution to efficiently encode relationship of each pair of nodes across two subgraphs, and efficiently derive in tensor space by utilizing the property of Kronecker product.
\item we experimentally validate the effectiveness of our method on both action recognition and matrix completion datasets, and report the state-of-the-art results.
\end{enumerate}

\begin{figure*}[!t]
  \centering
  \includegraphics[height=2.8in, width=6.8in]{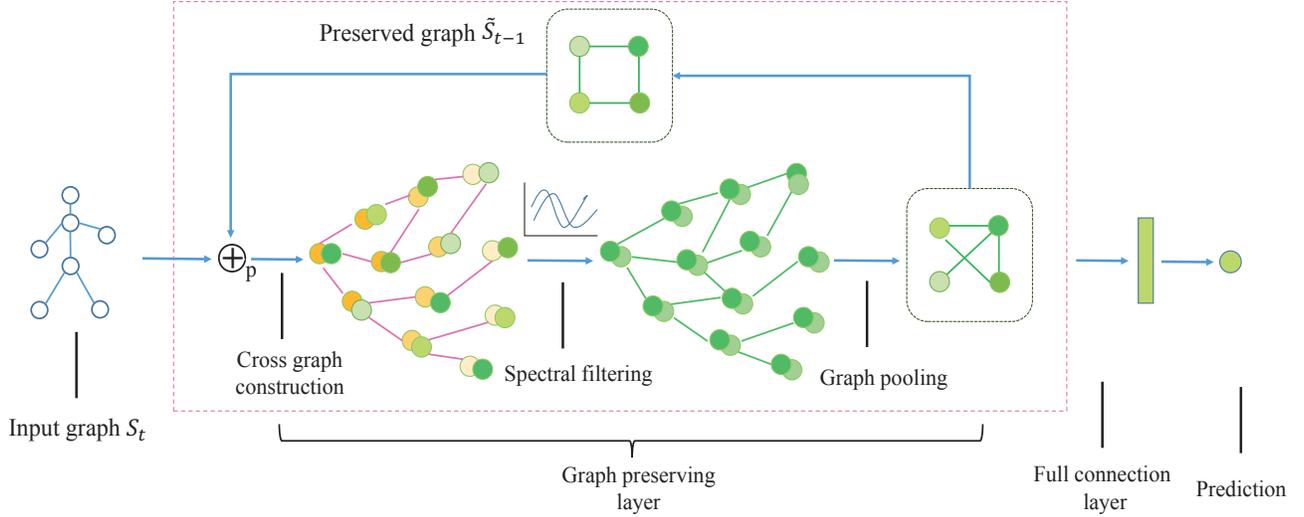}
  \caption{The architecture of TGCNN for action recognition on skeleton data. The key graph preserving layer is an recursive unit which aims to learn an optimal graph from all input factorized graphs.  }
  \label{framework}
\end{figure*}

\section{Preliminary of spectral filtering}
Let  $\mathbf{s} = [s_1, s_2, ..., s_n]^T$ be a signal where each element represents a vertex in a graph, and $\mathbf{A}$ be the corresponding adjacency matrix, then the spectral filtering on $\mathbf{s}$ can be formulated as:
  \begin{eqnarray}\label{eq_p1}
   \mathbf{\tilde{s}} = g_\theta(\mathbf{L})\mathbf{s} = g_\theta(\mathbf{U}\Lambda \mathbf{U}^T)\mathbf{s}=\mathbf{U}g_\theta(\Lambda )\mathbf{U}^T \mathbf{s}
 \end{eqnarray}
where
  \begin{eqnarray}\label{eq_p2}
           \mathbf{L} = \mathbf{I} - \mathbf{D}^{\frac{1}{2}} \mathbf{A} \mathbf{D}^{\frac{1}{2}} =  \mathbf{U}^T\Lambda \mathbf{U},  g_\theta(\Lambda)=\sum_{k=0}^{K-1}\theta_k\Lambda_k
  \end{eqnarray}
In Eqn.~\ref{eq_p2}, $\mathbf{D} \in \mathbb{R}^{n \times n}$ is a diagonal degree matrix  with the diagonal elements calculated as $D_{ii} = \sum_j{A_{ij}}$.

Furthermore, according to ~\cite{sandryhaila2013discrete}, $\mathbf{L}$ can be replaced by $\mathbf{A}$ as they have the same Fourier basis.
Then after substituting Eqn.~\ref{eq_p2} to Eqn.~\ref{eq_p1} and replacing  $\mathbf{L}$ with $\mathbf{A}$, spectral filtering can be rewritten as follows:
  \begin{eqnarray}\label{eq_p3}
 \mathbf{\tilde{s}}= \sum_{k=0}^{K-1} \theta_k \mathbf{A}^k \mathbf{s}
  \end{eqnarray}

\section{TGCNN architecture}
 The whole architecture of the proposed TGCNN is shown in Fig.~\ref{framework}, in which the key component  is the graph preserving layer.  In following subsections, we first describe the two  key operations of graph preserving layer, i.e. cross graph convolution and  graph pooling.  Then we show the whole recursive learning process of graph preserving layer.

\subsection{Cross graph convolution}
 Cross graph convolution consists of two main steps, i.e. cross graph construction and spectral filtering. Cross graph construction aims to build conjunctive graph between  two subgraphs, where each pair of nodes from  these two subgraphs are well modeled.  In cross graph construction process,  let $\mathbf{x} \in \mathbb{R}^{n_1}$ and $\mathbf{y} \in \mathbb{R}^{n_2}$ be two signals and $\mathbf{A}_x \in \mathbb{R}^{n_1 \times n_1}, \mathbf{A}_y \in \mathbb{R}^{n_2 \times n_2}$ be the corresponding adjacency matrices, then the conjunctive signal $\mathbf{S}=[\mathbf{s}_1, \mathbf{s}_2]$ is defined as follows:
\begin{eqnarray} \label{eq1}
   \nonumber   \mathbf{S} = [\mathbf{s}_1, \mathbf{s}_2] =  f(\mathbf{x}, \mathbf{y}) ~~~~~~~~~~~~~~~~~~~~~~~~~~~ \\
      ~~~~~~~   = \begin{bmatrix}
                                     \mathbf{x} \otimes (  \mathbf{1}^{n_2}), ~( \mathbf{1}^{n_1}) \otimes \mathbf{y}\\
 \end{bmatrix},
\end{eqnarray}
where
\begin{eqnarray}
 \nonumber    \mathbf{s}_1, \mathbf{s}_2 \in  \mathbb{R}^{n_1* n_2}, \mathbf{S} \in \mathbb{R}^{(n_1* n_2)\times 2}.
\end{eqnarray}
And according to spectral graph theory, the adjacency matrix of $\mathbf{S}$, denoted as $\mathbf{A} \in \mathbb{R}^{(n_1*n_2) \times (n_1*n_2)}$, should correspondingly describe the similarities between each pair of nodes in $\mathbf{S}$. In previous research, Kronecker sum operation is proposed which can be used for describing conjunctive  similarities of two given subgraphs:
\begin{equation} \label{eq2}
  \mathbf{A}= \mathbf{A}_x \oplus \mathbf{A}_y= \mathbf{A}_x \otimes \mathbf{I}^{n_2}+  \mathbf{I}^{n_1}\otimes\mathbf{A}_y.
\end{equation}
 However, due to the complexity of signals, the  conjunctive adjacency matrix generated by  classic Kronecker sum operation may not well fit $\mathbf{S}$. For this reason, we parameterize the  Kronecker sum operation expecting to learn a optimal conjunctive adjacency matrix from $\mathbf{A}_x, \mathbf{A}_y$. The new operation,  denoted as $\oplus_{p}$, is named as parameterized  Kronecker sum operation which is defined as follows:
 \begin{eqnarray} \label{eq3}
  \mathbf{A}= \mathbf{A}_x \oplus_{p} \mathbf{A}_y= \mathbf{A}_x \otimes \mathbf{I}_{\mathbf{\lambda}_1}+  \mathbf{I}_{\mathbf{\lambda}_2}\otimes\mathbf{A}_y, \\
  \text{where}~
    \mathbf{I}_{\mathbf{\lambda}} = diag(\mathbf{\lambda}),  \mathbf{\lambda}_1 \in \mathbb{R}^{n_2}, \mathbf{\lambda}_2 \in \mathbb{R}^{n_1}.
\end{eqnarray}
In Eqn.~\ref{eq3},  $\mathbf{\lambda_1},~\mathbf{\lambda_2}$ are both trainable vectors.

Then, we conduct cross graph convolution by applying spectral filtering on $\mathbf{S}$ and $\mathbf{A}$:
\begin{eqnarray}
     \nonumber  \mathbf{\tilde{S}} = \mathbf{H}_\theta(\mathbf{A}, \mathbf{S}) = \sum_{k=0}^{K}\theta_k \mathbf{A}^k \mathbf{S} ~~~~~~~~~\\
                \label{eq_4}       ~           = [\sum_{k=0}^{K}\theta_k\mathbf{A}^k \mathbf{s}_1, \sum_{k=0}^{K}\theta_k\mathbf{A}^k\mathbf{s}_2 ].~~~~~~
\end{eqnarray}
 In this process, the key step is to calculate the $k$-th order polynomials of the adjacency matrix. Formally, the $1$-st order polynomial on $\mathbf{s}_1$ can be calculated as follows:
\begin{eqnarray}
      \mathbf{A}\mathbf{s}_1 =   (\mathbf{A}_x \otimes \mathbf{I}_{\mathbf{\lambda}_1}+ \mathbf{I}_{\mathbf{\lambda}_2}\otimes\mathbf{A}_y)\mathbf{s}_1 ~~~~~~~~~~~~~~~~~~~~~~~~~~\\
                         \label{eq_6}           =  (\mathbf{A}_x \otimes \mathbf{I}_{\mathbf{\lambda}_1})\mathbf{s}_1 + (\mathbf{I}_{\mathbf{\lambda}_2}\otimes\mathbf{A}_y)\mathbf{s}_1~~~~~~~~~~~~~~~~~~~~~ \\
                          \label{eq_7}          = vec(\mathbf{I}_{\mathbf{\lambda}_1}mat(\mathbf{s}_1)\mathbf{A}_x^T)+vec(\mathbf{A}_y mat(\mathbf{s}_1) \mathbf{I}_{\mathbf{\lambda}_2})~~~~~
\end{eqnarray}
where $vec(\cdot)$ denotes the vectorization of a matrix by stacking its columns into a single column vector, and $mat(\cdot)$ is the reverse process of $vec(\cdot)$ which transforms a single column vector into a matrix. So, we also have the following equation:
\begin{eqnarray}
 mat(\mathbf{A}\mathbf{s}_1)=\mathbf{I}_{\mathbf{\lambda}_1}mat(\mathbf{s}_1)\mathbf{A}_x^T+\mathbf{A}_y mat(\mathbf{s}_1)\mathbf{I}_{\mathbf{\lambda}_2}.
\end{eqnarray}
In above equations, Eqn.~\ref{eq_6} is transformed to Eqn.~\ref{eq_7} by utilizing the property of Kronecker product. Then, the $2$-nd order polynomial on $\mathbf{s}_1$ can be further calculated as:
\begin{eqnarray}
  \mathbf{A}^2\mathbf{s}_1 = \mathbf{A}(\mathbf{A}\mathbf{s}_1) ~~~~~~~~~~~~~~~~~~~~~~~~~~~~~~~~~~~~~~~~~~~~~~\\
                                                  =(\mathbf{A}_x \otimes \mathbf{I}_{\mathbf{\lambda}_1}+\mathbf{I}_{\mathbf{\lambda}_2}\otimes\mathbf{A}_y)(\mathbf{A}\mathbf{s}_1)~~~~~~~~~~~~~~~~~~~~\\
                                                  = vec(\mathbf{I}_{\mathbf{\lambda}_1}mat(\mathbf{A}\mathbf{s}_1)\mathbf{A}_x^T+\mathbf{A}_y mat(\mathbf{A}\mathbf{s}_1)\mathbf{I}_{\mathbf{\lambda}_2})~~~~~
\end{eqnarray}
Based on the results of 1-st and 2-nd order polynomials,  the $K$-th order polynomial on $\mathbf{s}_1$ can be deduced, which can be written as:
\begin{eqnarray}
  \mathbf{A}^K\mathbf{s}_1 = \mathbf{A}(\mathbf{A}^{K-1}\mathbf{s}_1) ~~~~~~~~~~~~~~~~~~~~~~~~~~~~~~~~~~~~~~~~\\
                                                  = vec(\mathbf{I}_{\mathbf{\lambda}_1}mat(\mathbf{A}^{K-1}\mathbf{s}_1)\mathbf{A}_x^T+\mathbf{A}_y mat(\mathbf{A}^{K-1}\mathbf{s}_1)\mathbf{I}_{\mathbf{\lambda}_2}).
\end{eqnarray}
By utilizing the property of Kronecker product,  the graph convolution on $\mathbf{A}$, which needs to calculate polynomials on matrices in size $(n_1*n_2) \times (n_1*n_2)$,  is transformed to calculate the matrix product on two matrices in much smaller sizes,  i.e. $(n_1 \times n_1) $ and $(n_2 \times n_2) $ respectively. Correspondingly, the computational cost also decreases from $(n_1*n_2) \times (n_1*n_2)$ to $(n_1*n_1) + (n_2*n_2)$, which effectively relieves the computational burden.

Moreover, to further improve the computation efficiency,  the vectorization operation denoted as  $vec(\cdot)$ and its reverse operation  $mat(\cdot)$ need not be conducted frequently  by transforming Eqn.~\ref{eq_4} to the following format:
 \begin{eqnarray}
     \mathbf{\tilde{S}}= [vec(\sum_{k=0}^{K}\theta_kmat(\mathbf{A}^k \mathbf{s}_1)), vec(\sum_{k=0}^{K}\theta_kmat(\mathbf{A}^k\mathbf{s}_2) )]
 \end{eqnarray}

\subsection{Graph pooling}
As cross graph convolution results in a large graph of $n1*n2$ nodes from two factorized graphs where the  total number of nodes is $n1+n2$,  there may be some irrelevant nodes contributing little to action recognition. This part of nodes increase computation cost and also may degrade the performance. To reduce the disturbances of these irrelevant nodes as well as reduce the graph size, a projecting matrix is utilized to weight nodes so that an optimal subset of salient nodes can be effectively selected.

Formally, given the nodes denoted as  $\mathbf{S}$ and it's corresponding adjacency matrix $\mathbf{A}$, the graph pooling process can be described as follows:
 \begin{eqnarray}
p{(\mathbf{S})}= \mathbf{W}\mathbf{S},~~~~~\\
p{(\mathbf{A})}=\mathbf{W}\mathbf{A} \mathbf{W}^T.
 \end{eqnarray}
where $\mathbf{W}$ is the parameter to be solved. The larger the matrix element is, the more important the corresponding weighted node is for action recognition. This graph pooling operation not only further promotes the performance of TGCNN, but also avoids serious graph expansion during the recursive learning process  introduced in the following subsection in detail.

\subsection{Recursive learning process}
 The recursive learning process aims to transform the factorized graph modeling from processing all nodes at  one time to recurrent convolution on two subgraphs step by step, which effectively relieves the computational burden. Formally, let  $\mathbf{A}_1, \mathbf{A}_2, ...,  \mathbf{A}_T$ denote $T$ factorized subgraphs and $\mathbf{S}_1, \mathbf{S}_2, ...,  \mathbf{S}_T$ be their corresponding signals,  based on cross graph convolution and graph pooling, the recursive learning process can be formulated as follows:
\begin{eqnarray}
     \label{eq18}        \mathbf{S}^c_n = f(\mathbf{S}_n, \mathbf{\tilde{S}}_{n-1}),     ~~~~~~~ \\
     \label{eq19}        \mathbf{A}^c_n = \mathbf{A}_n \oplus_p \mathbf{\tilde{A}}_{n-1}, ~~~~~~ \\
     \label{eq20}        \mathbf{\tilde{S}}_n =  p(\mathbf{H}_\theta(\mathbf{A}^c_n, \mathbf{S}^c_n)), ~~~~ \\
     \label{eq21}        \mathbf{\tilde{A}}_{n} = p(\mathbf{A}^c_{n}), ~~~~~~~~~~~~~~\\
     \label{eq22}        \mathbf{O}_{n} =  \sigma_1(\mathbf{W}_{co}\mathbf{\tilde{S}}_n+\mathbf{b}_{co}).
 \end{eqnarray}
In above equations,  $\mathbf{A}^c_{n}$ and  $\mathbf{S}^c_n$ represent the conjunctively constructed adjacency matrix and signal respectively, which are  generated from the current input and previous preserved graphs.  $\mathbf{\tilde{A}}_{n}$ and  $\mathbf{\tilde{S}}_n$  denote the preserved adjacency matrix and the filtering signal at the $n$-th recursive step, and $\mathbf{O}_n$ is the $n$-th output feature. $\mathbf{W}_{co}$ and $\mathbf{b}_{co}$ are learnable variables for learning an output feature from the filtering signal, and $ \sigma_1(\cdot)$ is a non-linear activation function which is used for endowing flexibility to the graph preserving layer.

Eqn.~\ref{eq18}-\ref{eq22} cooperate to achieve global learning on all input factorized subgraphs. Given successive factorized graphs, the nodes between the preserved and input graphs are jointly modeled  through cross graph convolution (Eqn.~\ref{eq18}-\ref{eq20}). Then, based on this,  graph pooling is applied which acts as a memory unit to remember those salient nodes. These two operations are recursively conducted so that the learning process makes TGCNN be inherently deep as the previous input graphs are connected with current input one. And thus, the output features are able to describe the global graph structure.

\subsection{The loss functions}
For action recognition, the output features of graph preserving layer  are further passed through a full connection layer and a softmax layer. Then cross entropy loss is employed for TGCNN training, which can be defined as follows:
\begin{eqnarray} \label{eq5}
\nonumber E =-\sum_{i=1}^N\sum_{c=1}^C\tau(y_i,c) \times \log P(c|{\mathbf{S}_{i1}, \cdots, \mathbf{S}_{iT}})
\end{eqnarray}
in which
\begin{equation} \label{eq4}
   \nonumber   \tau(y_i,c)=
      \begin{cases}
         1,  &  \mbox{if} ~y_i=c; \\
         0,  &  \mbox{otherwise.}
       \end{cases}
\end{equation}
where $E$ denotes the cross entropy loss calculating the mean negative logarithm value of the prediction probability of the training samples, $N$ denotes the number of the training samples, $\mathbf{S}_{i1}, \cdots, \mathbf{S}_{iT}$ represent input factorized graphs of $i$-th training sample and $y_{i}$ is the corresponding label.

Besides action recognition, to test the generalization ability of the proposed cross graph convolution on graph structural data, we also conduct an extensive experiment on a matrix completion dataset~\cite{kalofolias2014matrix,monti2017geometric}. In matrix completion, only two factorized subgraphs are provided instead of multiple graphs. So, this task can be treated as a simplified application case of our TGCNN. As the algorithm in~\cite{monti2017geometric} achieves the current best performance, for a fair comparison, we embed our cross graph convolution into the framework of~\cite{monti2017geometric} and employ the same loss function as~\cite{monti2017geometric}, which is commonly employed in matrix completion algorithms.

\section{Experiments}\label{Experiments}
 We evaluate  our TGCNN on two action recognition datasets named NTU RGB+D (NTU) dataset~\cite{shahroudy2016ntu} and  Large Scale Combined (LSC) dataset~\cite{zhanglarge}. Besides, to test the generalization ability to graph structural data, we also conduct an extensive experiment on a matrix completion dataset named Synthetic ‘Netflix’ dataset~\cite{kalofolias2014matrix}, in which two factorized graphs are provided describing the relationship among users and items respectively. In the following subsections, we firstly introduce these three datasets, then we show the implemental details including the preprocessing of action recognition datasets. Finally, we compare the experimental results with the state-of-the-art methods.

\subsection{Datasets}\label{Dataset}
In NTU dataset, there are 56880 RGB+D video samples executed by 40 different human subjects whose ages are in the range from 10 to 35.  Three synchronous  Microsoft Kinect v2 sensors are used for collecting various modalities of signals from three different horizontal angles, where the modalities include RGB videos, depth sequences, skeleton data and infrared frames. Specifically, for skeleton data, the human skeleton is represented by 3D locations of 25 major body joints. This dataset is great challenging due to its large amount of samples, multiple view points and  intra-class variations.

The LSC dataset is an integrated dataset created by combing nine existing public datasets. In this dataset, there are 4953 video sequences containing red, green and blue (RGB) video and depth information. These sequences contain 94 action classes which are performed by 107 subjects in total.  As these video sequences come from different individual datasets, the variations with respect to subjects, performing manners and backgrounds are very large. Moreover,  there is large difference among the number of samples of each action. All these factors, i.e. the large size, the large variations and the data imbalance for each class, make this dataset challenging for recognition.

Synthetic ‘Netflix’ dataset is  frequently used in matrix completion task to evaluate different algorithms. In this dataset, the row axis represents different item (e.g. movie) while  the column represents different users.  Thus the value of each element  shows whether a user would like an item or not.  When creating this dataset,  the matrix  is generated by satisfying certain assumptions, e.g. low rank property and smoothness along rows and columns. Thus,  there is strong communities structure in the generated user and item graphs. The advantage of  this dataset is that it enables the behaviours of different algorithms  be well studied  in controlled settings.


\subsection{Preprocessing on skeleton data}\label{preprocessing}
 The preprocessing of skeleton data aims to eliminate noise and also make the model be robust to different kinds of variations, e.g. body orientation variation and body scale variation. This process is done by the following three steps:

(i) The action sequences are first split to a fixed number of subsequences, and then one frame is chosen from each subsequence so that the generated sequences contain the same number of frames.

(ii) The skeletons are randomly scaled with different factors ranging in [0.95, 1.05] so that the adaptive scaling capacity of the model can be improved.

(iii)  During training stage, the skeletons randomly are rotated  along $x,y$ and $z$ axis with angles ranging in [-45, 45], which makes the model be robust to orientation variation.

\subsection{Experiment on NTU dataset}\label{NTU}
The experiments on NTU dataset are conducted following two different protocols,  named cross subject  and cross view protocols respectively, in~\cite{shahroudy2016ntu}. For cross subject protocol,  samples are  split to training and testing sets according to subjects' ID numbers. Under this protocol,  the split training and testing sets contain 40320 and 16560 samples respectively where the samples in each set are conducted by 20 subjects. For cross view protocol, there are 37920 and 18960 samples in training and testing sets respectively. Among them, samples in training set are captured by  cameras $\#$2 and $\#$3 while the samples in testing set are captured by camera $\#$1.

The main parameters in TGCNN are the polynomial order denoted as $K$, the numbers of preserved nodes and the dimension of  output features. For both protocols, K is set to be 2, the numbers of preserved nodes are both 50 and the dimension of output feature is 128.  Table~\ref{Label_NTU} shows the comparison results on NTU dataset. The proposed framework is compared with various the-state-of-the-art methods, including different kinds of recurrent neural networks (RNNs),  hierarchical bidirectional recurrent neural networks (HBRNN)~\cite{du2015hierarchical},  part-aware LSTM (P-LSTM)~\cite{shahroudy2016ntu}, spatio-temporal LSTM (ST-LSTM)~\cite{liu2016spatio}, , and geometric features LSTM (GF-LSTM)~\cite{zhang2017geometric}. For both protocols, our algorithm achieves the best performance.

\begin{table}
\renewcommand{\arraystretch}{1.35}
\begin{center}
\scalebox{1.1}{
\begin{tabular}{|c|c|c|}
\hline
Method                                                                                      & \tabincell{c}{Cross Subject\\ Accuracy ($\%$)}   & \tabincell{c}{Cross View\\ Accuracy($\%$) }  \\
\hline
\tabincell{c}{Lie group~\cite{vemulapalli2014human}}    & 50.08                                                                         & 52.76\\
\hline
\tabincell{c}{HBRNN~\cite{du2015hierarchical}}             & 59.07                                                                         & 63.97\\
\hline
\tabincell{c}{LieNet~\cite{huang2017deep}}                      &  61.37                                                                        & 66.95\\
\hline
\tabincell{c}{Deep LSTM~\cite{shahroudy2016ntu}}         & 60.69                                                                         & 67.29\\
\hline
\tabincell{c}{P-LSTM~\cite{shahroudy2016ntu}}               & 62.93                                                                        & 70.27\\
\hline
\tabincell{c}{ST-LSTM~\cite{liu2016spatio}}                     & 69.20                                                                         & 77.70\\
\hline
\tabincell{c}{GF-LSTM~\cite{zhang2017geometric}}        & 70.26                                                                         & 82.39\\
\hline
 TGCNN                                                                                    & \textbf{71.4}                                                            &\textbf{82.9}\\
\hline
\end{tabular}}
\end{center}
\caption{The comparisons on NTU dataset.}
\label{Label_NTU}
\end{table}

\subsection{Experiment on Large Scale Combined dataset}\label{LSC}
We conduct experiments on LSC dataset by following two different protocols employed in~\cite{zhanglarge}. For the first protocol named  Random Cross Sample (RCSam) using data of 88 action classes,  half of the samples of each class are randomly selected as training data while the rests are used as testing data.  For the second protocol named  Random Cross subject (RCSub) using data of  88 action classes,  half of the subjects are randomly selected as training data and the rest subjects are used as testing data. In both protocols, only skeleton data are used for recognition. Due to the imbalance of samples in each class,  the values of  precision and  recall are employed for evaluating the performance instead of accuracy.

\begin{table}[t]
\renewcommand{\arraystretch}{1.5}
\begin{center}
\scalebox{1}{
{\begin{tabular}{|c|c|c|c|}
\hline
Protocol & Method & ~~~~\tabincell{c}{Precision\\($\%$)}~~~~ & ~~\tabincell{c}{Recall \\ ($\%$)}~~ \\
\hline
\multirow{5}{*}{RCSam}&\tabincell{c}{HON4D ~\cite{oreifej2013hon4d}}  & 84.6& 84.1 \\
 \cline{2-4}{} &\tabincell{c}{Dynamic skeleton~\cite{zhanglarge}} & 85.9 &   85.6 \\
 \cline{2-4}{} &\tabincell{c}{P-LSTM~\cite{shahroudy2016ntu}} & 84.2 &   84.9 \\
 \cline{2-4}{} &TGCNN   &   \textbf{86.6}  & 82.9 \\ \hline
\multirow{5}{*}{RCSub}&\tabincell{c}{HON4D~\cite{oreifej2013hon4d}}& 63.1 & 59.3 \\
 \cline{2-4}{} &\tabincell{c}{Dynamic skeleton ~\cite{zhanglarge}} & 74.5 &  73.7 \\
\cline{2-4}{} &\tabincell{c}{P-LSTM~\cite{shahroudy2016ntu}} & 76.3 &  74.6 \\
 \cline{2-4}{} &TGCNN  & \textbf{83.1}  & \textbf{76.5} \\ \hline
\end{tabular}}}
\end{center}
\caption{The comparisons on LSC dataset following RCSam and RCSub protocols.}
\label{LSC_result}
\end{table}

The parameter settings for both RCSam and RCsub protocols are the same: the numbers of nodes in preserving layer are set to be 30 while the dimension of output feature is set to be 80, and $K$  is set to be 2.  The comparisons on LSC dataset are shown in Table~\ref{LSC_result}. Except the recall value in RCSam protocol, our algorithm outperforms the previous state-of-the-art methods.

\subsection{Experiment on Synthetic ‘Netflix’ dataset}\label{Synthetic}

\begin{table}[t]
\renewcommand{\arraystretch}{1.5}
\begin{center}
\scalebox{1.1}{
\begin{tabular}{c}
\hline
\hline
~~~~~~Methods    ~~ ~~~~~~~~~~~~~  Complexity    ~~~~~~~~~~~~~~~~                    RMSE      \\
\hline
~~~~~~GMC~\cite{kalofolias2014matrix}         ~~~~~~~~~~~~~~~~~~~~     mn              ~~~~~~~~~~~~~~~~~~~~~        0.3693          \\
~~~~GRALS~\cite{rao2015collaborative}       ~~~~~~~~~~~~~~~~      m~+~n            ~~~~~~~~~~~~~~~~~~~           0.0114       \\
~~~~RGCNN~\cite{monti2017geometric}    ~~~~~~~~~~~~~~~~~    mn              ~~~~~~~~~~~~~~~~~~~~~       0.0053  \\
~~~sRGCNN~\cite{monti2017geometric}   ~~~~~~~~~~~~~~~     m~+~n       ~~~~~~~~~~~~~~~~~~~        0.0106      \\
~~~~~~TGCNN   ~~~~~~~~~~~~~~~~~~     mm+nn          ~~~~~~~~~~~~~~~~~    \textbf{ 0.0042}    \\
\hline
\end{tabular}}
\end{center}
\caption{The comparisons on Synthetic ‘Netflix’ dataset.}
\label{Label_Synthetic}
\end{table}

\begin{table}[t]
\renewcommand{\arraystretch}{1.3}
\begin{center}
\scalebox{1.1}{
\begin{tabular}{|c|c|c|c|c|}
 \hline
 \multirow{2}{*}{Architecture} &  \multicolumn{2}{c|}{\tabincell{c}{LSC dataset \\ (RCSam protocol)}}&  \multicolumn{2}{c|}{\tabincell{c}{LSC dataset \\ (RCsub protocol)}}  \\
 \cline{2-5}  & Precision &  Recall  & Precision   & Recall    \\
 \hline
IGCNN & $80.4$ & $77.7$  & $80.7$ & $73.2$   \\
 \hline
 TGCNN & $\textbf{86.6}$ & $\textbf{82.9}$ & $\textbf{83.1}$ &  $\textbf{76.5}$  \\
 \hline
 \end{tabular}}
 \end{center}
 \caption{Factorized graph convolution vs isolate graph convolution.}
\label{FIGCNN}
\end{table}

We follow the protocol employed in~\cite{monti2017geometric} to evaluate the performance of our TGCNN on Synthetic ‘Netflix’ dataset.  Under this protocol, a part of  chosen values of users are first eliminated, and  the task of algorithm  is to recovering the missing values of this matrix based on the given  fraction of  entries. At last, root mean squared (RMS) error is employed to evaluate the difference between the recovered matrix and the ground truth. The smaller value of RMS error means the  better performance.

The results of different matrix completion methods are reported in Table~\ref{Label_Synthetic}, along with their theoretical complexities. Algorithms including geometric matrix completion
(GMC)~\cite{kalofolias2014matrix}, recurrent graph CNN (RGCNN)~\cite{monti2017geometric}, separable recurrent graph CNN (sRGCNN)~\cite{monti2017geometric} and graph regularized alternating least squares (GRALS)~\cite{rao2015collaborative}, are compared with our TGCNN. As it is shown,  our TGCNN model achieves the best accuracy which demonstrates the generalization ability of TGCNN for graph structural data.

\begin{table}[t]
\renewcommand{\arraystretch}{1.3}
\begin{center}
\scalebox{1.1}{
\begin{tabular}{|c|c|c|c|}
 \hline
 \multirow{2}{*}{Architecture}&  \multicolumn{2}{c|}{ \tabincell{c}{ LSC dataset \\  (RCSam protocol)}} &    \multirow{2}{*}{  \tabincell{c}{ Synthetic \\  ‘Netflix’ dataset}} \\
 \cline{2-3}  & Precision & Recall &   \\
 \hline
Kronecker sum & $ 85.2 $ & $ 82.4$ & $ 0.0049$   \\
 \hline
 P-Kronecker sum & $\textbf{86.6}$ & $\textbf{82.9}$& $\textbf{0.0042}$   \\
 \hline
 \end{tabular}}
 \end{center}
 \caption{Parameterized Kronecker  (P-Kronecker) sum vs  classic Kronecker sum on LSC dataset and Synthetic ‘Netflix’ dataset.}
\label{PclassicKron}
\end{table}

\subsection{Analysis of TGCNN}

\begin{table}[t]
\renewcommand{\arraystretch}{1.5}
\begin{center}
\scalebox{1.1}{
\begin{tabular}{c}
\hline
\hline
~~K~~     ~~~~~ ~   \tabincell{c}{ Precision on LSC   \\ dataset ( $\%$)}  ~~~~~ ~     \tabincell{c}{ Recall on LSC   \\ dataset ( $\%$)}     \\     
\hline
~~~~~~2              ~~~~~~~~~~~~~~~~~~~                        $\textbf{83.1}$             ~~~~~~~~~~~~~~~~~~~   $76.5$       ~~~~~~~~    \\     
~~~~~~3             ~~~~~~~~~~~~~~~~~~~                         $81.3$           ~~~~~~~~~~~~~~~~~~~   $76.4$      ~~~~~~~~           \\   
~~~~~~4             ~~~~~~~~~~~~~~~~~~~                         $82.8$           ~~~~~~~~~~~~~~~~~~~   $76.3$     ~~~~~~~~                       \\   
~~~~~~5             ~~~~~~~~~~~~~~~~~~~                         $82.5$           ~~~~~~~~~~~~~~~~~~~   $\textbf{77.7}$       ~~~~~~~~                       \\
\hline
\end{tabular}}
\end{center}
\caption{Performance of TGCNN under different orders of polynomials on LSC datasets following RCSub protocol.}
\label{Label_Analysis}
\end{table}

As  TGCNN achieves promising performance, it is meaningful to verify how much the novel proposed operations, e.g. cross graph convolution and parameterized Kronecker sum, improve the performance of the network, and also how the parameter setting influences the  result. For these purposes, several additional experiments are respectively conducted which are listed as follows:
\begin{enumerate}
\item[(1)] TGCNN vs isolate graph CNN (IGCNN). To evaluate the effectiveness of the proposed recursive process, we conducted additional experiments on LSC datsets  comparing the results of TGCNN with IGCNN, where in IGCNN the graph filtering is only conducted  on each isolate graph (Table~\ref{FIGCNN}).
\item[(2)] Parameterized Kronecker sum  vs classic Kronecker sum. To see how much improvement the parameterized Kronecker sum operation brings, we conducted experiments on both LSC and  Synthetic ‘Netflix’ dataset to compare the performance (Table~\ref{PclassicKron}).
\item[(3)] Setting different orders of polynomials. We conduct additional experiments on LSC dataset following RCSub protocol to see how different  polynomial orders influent the performance  (Table~\ref{Label_Analysis}).
\end{enumerate}
From the results we can have the following observations:
\begin{enumerate}
\item[(i)] TGCNN outperforms IGCNN which verifies the effectiveness of the recursive learning process, which globally learns the additional cross graph relationship comparing to IGCNN.
\item[(ii)] The parameterized Kronecker sum operation further promotes the performance comparing to the classic one. This indicates that through parameterized Kronecker sum operation, the constructed conjunctive cross graph better fits the corresponding conjunctive signal.
\item[(iii)] The value of polynomial order influences the performance of graph filtering.  On LSC dataset, the best precision is achieved by setting K to be 2 while the best recall is achieved when K is set to be 5.
\end{enumerate}

\section{Conclusion}
In this paper, we proposed a novel framework named  TGCNN  to globally model those subgraphs factorized from a large graph. For this purpose, we  propose a  recursive learning process  on graph by specifically designing a novel graph preserving layer. Serving as a memory unit, this graph preserving layer memorizes those salient nodes of successively input graphs, where the  memory function is achieved through applying  novelly designed cross graph convolution and graph pooling. Specifically, cross graph convolution well models the relationship of each pair of nodes across graph and can be efficiently conducted. Besides, the proposed parameterized Kronecker product learns an optimal conjunctive adjacency matrix which further promotes the performance.  Comprehensive experiments conducted on  action recognition and matrix completion datasets verify the competitive performance of our TGCNN.


%

%
%
%
%
%

\ifCLASSOPTIONcaptionsoff
  \newpage
\fi

\end{document}